\documentclass[conference]{IEEEtran}
\IEEEoverridecommandlockouts
\usepackage{cite}
\usepackage{amsmath,amssymb,amsfonts}
\usepackage{algorithmic}
\usepackage{graphicx}
\usepackage{textcomp}
\usepackage{xcolor}
\usepackage{todonotes}
\usepackage{booktabs}
\usepackage{multirow}
\usepackage{siunitx}
\usepackage{graphicx}
\usepackage{subcaption}
\usepackage{pgfplots}
\usepackage{tabularx}
\usepackage{makecell}
\usepackage{multirow}

\makeatletter
\let\NAT@parse\undefined
\makeatother
\usepackage[hidelinks]{hyperref}
\usepackage[capitalize]{cleveref}
\usepackage{cite}

\pgfplotsset{compat=1.18}
\def\BibTeX{{\rm B\kern-.05em{\sc i\kern-.025em b}\kern-.08em
    T\kern-.1667em\lower.7ex\hbox{E}\kern-.125emX}}

\DeclareMathOperator*{\argmin}{arg\,min}
    
\begin{document}

\title{The Quadruped Soft Tail: Compliant Grasping and Swabbing for Contamination Surveys in Harsh Environments}

\author{Harald Minde Hansen$^{ \dagger 1}$, Nandita Gallacher$^{\dagger 1}$,\\ Kristin Y. Pettersen$^{2}$, Jan Tommy Gravdahl$^{2}$, and Mario di Castro$^{1}$
\thanks{$\dagger$ H.M. Hansen and N. Gallacher contributed equally to this work and should be considered co-first authors.}
\thanks{$^{1}${Mechatronics, Robotics and Operation Section, European Organization for Nuclear Research (CERN), Meyrin, Switzerland {\tt\small [{\href{mailto:harald.minde.hansen@cern.ch}{harald.minde.hansen}, \href{mailto:nandita.gallacher@cern.ch}{nandita.gallacher}, \href{mailto:mario.di.castro@cern.ch}{mario.di.castro}}]@cern.ch}}}%
\thanks{$^{2}${Department of Engineering Cybernetics, Norwegian University of Science and Technology (NTNU), Trondheim, Norway
{\tt\small [{\href{mailto:kristin.y.pettersen@ntnu.no}{kristin.y.pettersen}, \href{mailto:jan.tommy.gravdahl@ntnu.no}{jan.tommy.gravdahl}}]@ntnu.no}}}
}


\maketitle

\begin{abstract}
Beryllium contamination surveys in radioactive areas are challenging for robots in environments cluttered with cables and electronics. To address this problem, we have developed a novel quadruped system augmentation: A lightweight, soft, and compliant tendon-actuated robotic tail mounted on a quadruped robot. The tail features a hollow, flexible backbone and a tendon-actuated soft gripper that enables the robot to pick up sampling tissues, swab contaminated surfaces, and release the tissues at designated collection locations for subsequent beryllium analysis. To enable intuitive teleoperation, a closed-form kinematic model and a singularity-robust task-space controller are developed. Experimental results demonstrate that gripper actuation has a negligible effect on robot shape, while common-mode tendon actuation provides an effective mechanism for stiffness modulation and preload control. Furthermore, experimental validation indicates that the proposed kinematic model provides a suitable basis for real-time task-space control. The proposed system combines the agility of legged locomotion with the compliance of soft robotic manipulation, enabling the complete contamination-survey procedure to be performed without human exposure (see accompanying video). While motivated by beryllium contamination surveys at CERN, the proposed quadruped soft-tail concept is broadly applicable to legged robots operating in cluttered, confined, or hazardous environments where conventional rigid-link manipulators are undesirable.
\end{abstract}


\section{Introduction}

In the facilities of the European Organization for Nuclear Research (CERN), elementary particles are accelerated and collided at nearly the speed of light to study the fundamental properties of matter and their governing forces. This is possible through a complex of underground tunnels with superconducting magnetic and radio-frequency electric fields to concentrate and accelerate the particle beams before colliding them in experimental areas. The high energy levels of the particle beams cause the accelerator areas to become highly radioactive. This complicates tasks such as inspection, maintenance, and repair, necessitating robotic solutions to limit human radiation exposure.

\begin{figure}
\centerline{\includegraphics[scale=.07]{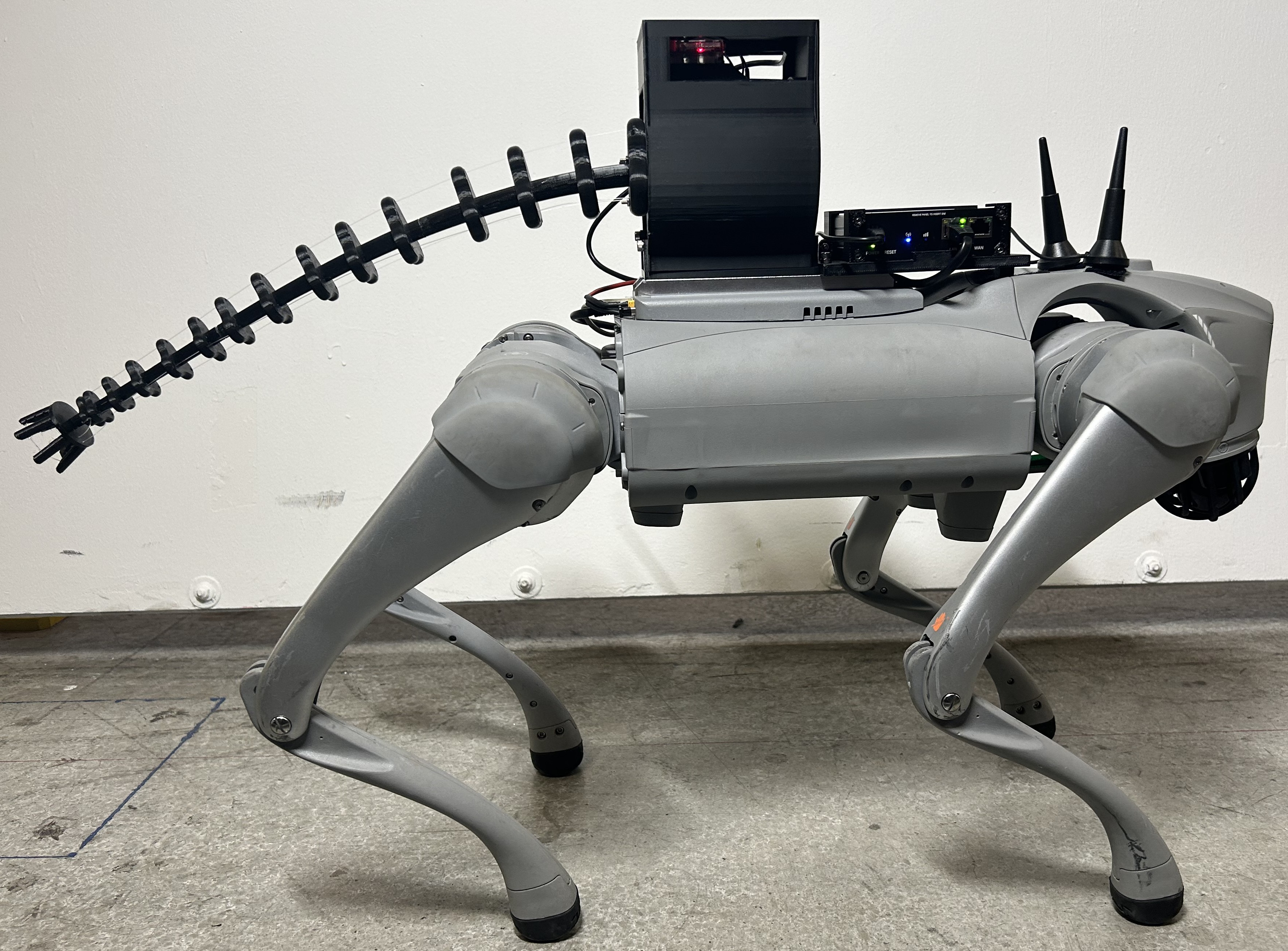}}
\caption{Quadruped soft tail system with tendon-actuated gripper.}
\label{fig:overview}
\vspace{-6mm}
\end{figure}

Many CERN areas present terrain that is difficult to navigate with standard wheeled robotic platforms. For example, the floor of the North Area experimental zone of the accelerator complex is heavily cluttered with electrical cables, cooling pipes and other obstacles. These areas cannot be accessed by wheeled platforms due to the height, gradient, and dynamic nature of these obstacles. Tracked platforms can access these areas in some cases, but the tracks can damage or disconnect cables and pipes. Additionally, in many areas, intervention missions also require the robot to travel underneath obstacles.

Quadruped robots have been deployed in interventions in these types of areas as a safer alternative. Since legged robots move by making and breaking contact with the environment, the feet can climb the non-rigid cables and find rigid contact in cable-heavy environments with no impact on the cables and pipes themselves. Through multiple interventions, quadruped robots have proved to be highly successful in these areas where no robot had accessed before. Successful interventions using quadrupeds in the North Area for detecting water leaks, machine inspection, and fault diagnosis have shown the flexibility and suitability of these robots to this difficult terrain.



To perform intervention tasks beyond inspection, the quadruped must be outfitted with tools. A number of robotic manipulators have been installed on quadrupeds in literature. Some of these are bio-inspired tails, and their function can range from locomotion assistance \cite{ibrahim2020, norby2021, liu2021, liu2025a, an2022, de2015, kohut2013}, to stabilization \cite{buckley2023, tang2023, rone2019}, tool carrying \cite{bellicoso2019}, and proprioceptive sensing \cite{yang2023}. Current literature on quadruped tails focuses extensively on rigid-link manipulators designed to improve maneuverability and stability, although some flexible tails have been implemented as compliant, compact, lower mass solutions \cite{liu2025, buckley2023, simon2018, liu2025a, rone2014, rone2018, saab2019, butt2021}. Some designs feature discs \cite{rone2014} and tendon-actuated variable stiffness tails \cite{buckley2023}, but for the sole function of improving locomotion without offering additional manipulation functionality.
Current literature on soft robotic grippers focuses primarily on factory automation \cite{dzedzickis2024, paderon2025}, agricultural \cite{zhang2025, goulart2023, bluiminck2023, chen2022}, and biomedical \cite{raja2024, shahid2024} applications, as their compliance makes them useful for grasping objects that are fragile or organic in shape, like produce. Some gripper designs aim to address issues with picking up objects that lie flat on surfaces, like paper and wipes. Designs include grippers with multiple actuating fingers \cite{lin2025,babin2019}, crawling fingers \cite{ko2020}, rolling fingers \cite{jiang2025}, and vibrating fingers \cite{nahum2022}, and require complex, high mass actuation systems and electronics. Wiping tools for radiation-contaminated environments have also been developed \cite{kearney2023}. While the compressible, passively compliant sections incorporated in the gripper designs improve wipe-surface contact, the grippers are solely capable of wiping. A compact, low-mass, versatile soft gripper solution is therefore desirable to address the limitations of these designs.




The work presented in this paper addresses the challenge of performing beryllium contamination surveys in cluttered radioactive environments. To this end, we propose a novel soft quadruped tail solution with a tendon-actuated soft gripper. The system combines legged mobility, compliant manipulation, and intuitive teleoperation in a single platform. In particular, the system is designed to accurately pick and place tissues, with the gripper capable of handling the worst-case scenario of tissues lying flat and unfolded on a surface. In addition, a closed-form kinematic model and a singularity-robust task-space controller are developed to enable real-time control of the soft tail.

The paper is organized as follows: \cref{sec:system_overview} describes the requirements for the intervention and the proposed robot solution, and \cref{sec:soft_gripper} details the soft gripper design. A kinematic model of the robot is outlined in \cref{sec:kinematic_model}, before a task-space kinematic controller with compliance is presented in \cref{sec:control}. The kinematic model and its assumptions are assessed through experiments in \cref{sec:experimental_results}. Lastly, \cref{sec:conclusion} concludes the paper and proposes future work.

\section{System Overview}\label{sec:system_overview} 

In this section, we present the requirements for the requested intervention, as well as our proposed solution. We present a compliant and flexible soft quadruped tail and its design tailored for this particular intervention.

\subsection{Task overview}

The radiation protection (RP) group at CERN has requested beryllium contamination surveys to be performed in some of the experimental areas. The requested procedure is as follows: Picking up a cotton tissue from a designated pick-up point, gently swabbing the walls with the tissue, and then releasing the tissue at a designated release point. This process should be repeated with multiple tissues at different locations. From the release point, the tissues can be examined by the RP group.

The maximum height of the quadruped with installed equipment in previous interventions is $616$ mm. We will avoid exceeding this height to ensure the system can still reach all areas. For this reason, a manipulator system installed on top of the quadruped must be able to fold to satisfy this constraint, such that it does not exceed a height of $183$ mm from the mounting plate. Moreover, to minimize the impact of the quadruped dynamics, we seek a light-weight solution that keeps the center of gravity of the quadruped as close as possible to that of the unmodified quadruped. The manipulator should also be able to bend its tip towards the floor as well as extend the total length of the quadruped in at least one horizontal direction for swabbing the walls. In order to avoid damage to walls or any equipment, the manipulator should be able to compliantly interact with the environment. 



\subsection{Proposed solution}

The soft robot developed for the Continuous Tendon-Actuated Manipulator (CTAM) project at CERN \cite{hansen2026} meets the above-mentioned requirements, ensuring a light-weight, flexible, and compliant solution for gentle interaction with its surroundings, where CTAM is installed as a quadruped tail, as depicted in \cref{fig:overview}. To this end, we have chosen the physical robot parameters such as backbone length, taper angle, and tendon routing following the methodology in \cite{hansen2026} to obtain suitable Cosserat-rod predicted curvature profiles, such that the robot tip reaches the floor and walls at a perpendicular tip tangential angle to the surfaces. Moreover, we have developed tendon-actuated soft grippers that are able to pick up tissues, swab the walls, and release the tissues. The full system is controlled from the computer integrated into the quadruped and teleoperated over WiFi.

For the quadruped integration, the backbone is made hollow with a fixed inner diameter such that the tendon actuating the gripper can be routed through the backbone center, and the disc sizes allow the routing of a 4.5 mm wide endoscopic camera through the discs and base of the gripper. At 405 mm, the length of CTAM is sufficient to allow the gripper to reach both a tissue laid flat on the floor and to reach past the rear legs of the quadruped to make contact with the curved tunnel wall. The system mass, including CTAM, the actuation system, the endoscopic camera, the gripper, the cover, electronics, and cabling, totals $902.8$ g. 

\section{Soft Gripper}\label{sec:soft_gripper}

In this section, we present the tendon-actuated soft gripper design. To simplify the fabrication process and design iterations, the gripper is printed in TPU 95A, the same material used for the CTAM backbone.






\subsection{Design}

\begin{figure}
\centerline{\includegraphics[scale=.17]{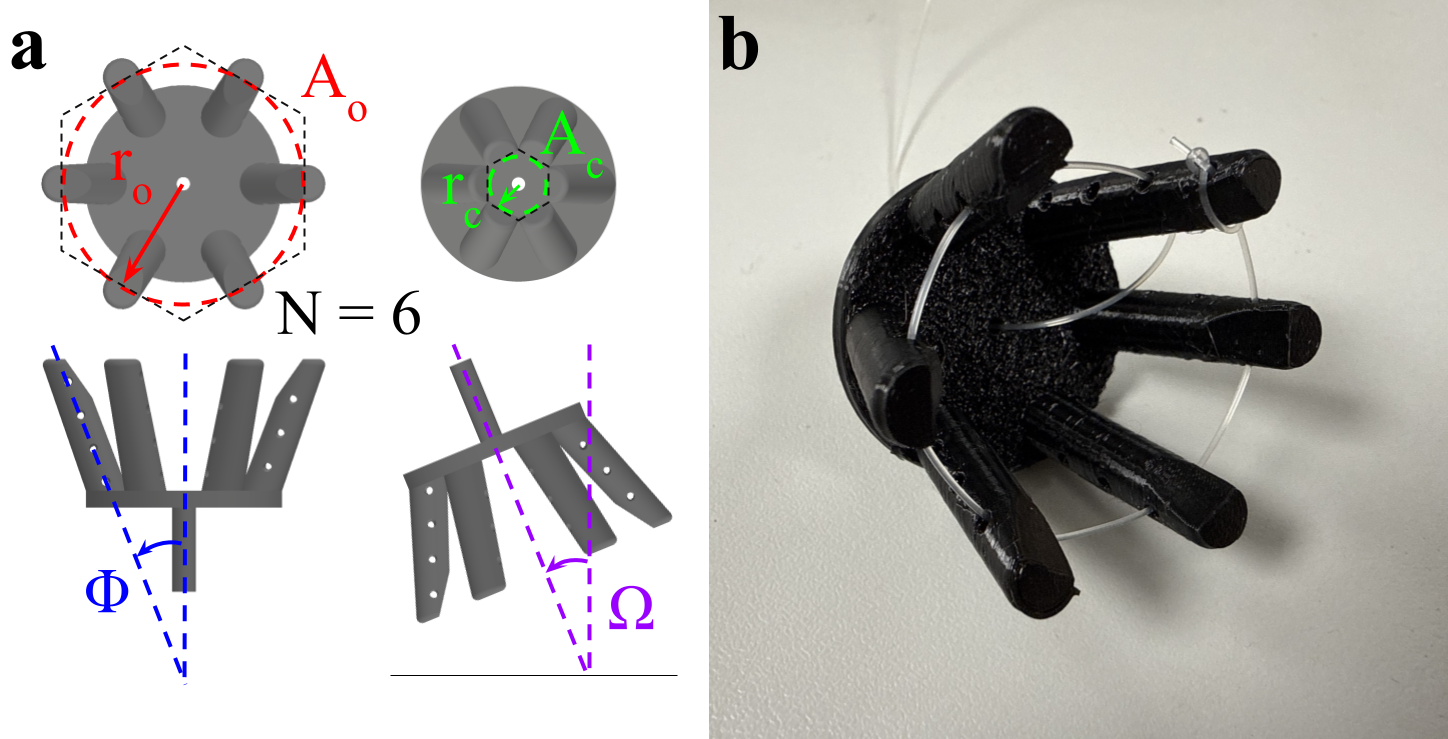}}
\caption{a) Diagram of gripper design parameters. b) Tendon-actuated soft gripper with $N = 6$ fingers at $\Phi = 20^\circ$.}
\label{fig:soft_gripper}
\vspace{-4mm}
\end{figure}

\cref{fig:soft_gripper}a shows the parametrized soft gripper design, with number of fingers, $N$, finger angle, $\Phi$, relative to the centerline axis of the gripper, effective area of contact when the gripper is open, $A_o$, and effective area of contact when the gripper is closed, $A_c$. The areas of contact are approximated as the circles inscribed within the N-sided polygon formed by the flat interior edges of each finger. \cref{fig:soft_gripper}b shows the gripper used in the accompanying submission video (see \hyperref[appendix]{Appendix}), with parameters $N = 6$ and $\Phi = 20^\circ$. The actuating tendon is routed helically, and pulling the tendon causes all the fingers to move towards the center to close the gripper. Releasing the tension causes the gripper to open, as the TPU retains its original shape. At the tip of each finger on its interior side is a flat edge, angled at 20 degrees about the centerline axis. 


\subsection{Experimental Testing}

\cref{tab:helical_gripper_force} presents an overview of the tendon tensions required to obtain contact between all gripper fingers and time for fingers to rebound from fully closed to to fully open over a variation of gripper infill densities and number of tendon traverses through each finger. The data demonstrate a clear increase in required actuation force with increasing infill density and number of tendon traverses. 
Qualitative observations also suggest that configurations with a greater number of tendon traverses exhibited a greater tendency toward residual deformation after unloading, potentially due to increased internal friction, localized stress concentrations, and prolonged distributed loading along the finger structure. Thus, the mean rebound time for more than 1 traverse is omitted from the table since the grippers never returned to their original state. As the infill density increases, the mean rebound time decreases, with a sharp decline between 40\% and 60\%. The higher infill prototypes showed faster recovery time, as the higher infill fingers exerted greater force to return to the relaxed open state. After reaching 60\%, the mean rebound time deviated only slightly, likely due to friction differences caused by printing defects.
For light object grasping, lower infill densities and a single tendon traverse appear to provide the most favorable trade-off between actuation force and shape recovery. For the fastest gripper shape recovery, an infill density of 60\% or 80\% is favorable, although all infill densities tested are suitable for the intervention task considering the time between releasing and grasping wipes is greater than 30 seconds. Since infill density was not found to have an impact on grasping functionality, this parameter is not considered in pick-and-place testing.


\begin{table}[ht]
\centering
\caption{Mean actuation force required to fully close the $N = 6$, $\Phi = 20^\circ$ soft gripper and mean rebound time for 1 traverse ($\pm$ one standard deviation, $n=5$).}
\label{tab:helical_gripper_force}
\renewcommand{\arraystretch}{1.3}
\begin{tabular}{c c c c}
\toprule
\makecell{Infill \\ Density (\%)} &
\makecell{1 Traverse \\ (N)} &
\makecell{2 Traverses \\ (N)} &
\makecell{Mean Rebound \\ Time (s)} \\
\midrule
20  & $5.91 \pm 1.04$  & $7.00 \pm 0.31$  & $10.408 \pm 3.057$ \\
40  & $8.44 \pm 1.18$  & $9.37 \pm 0.89$  & $9.156 \pm 2.960$ \\
60  & $11.54 \pm 0.62$ & $12.24 \pm 0.88$ & $5.434 \pm 0.843$ \\
80  & $13.31 \pm 1.30$ & $14.12 \pm 1.55$ & $5.378 \pm 0.489$ \\
100 & $14.46 \pm 1.95$ & $15.15 \pm 1.36$ & $5.806 \pm 1.101$ \\
\bottomrule
\end{tabular}
\end{table}

\begin{figure}
\centerline{\includegraphics[scale=.17]{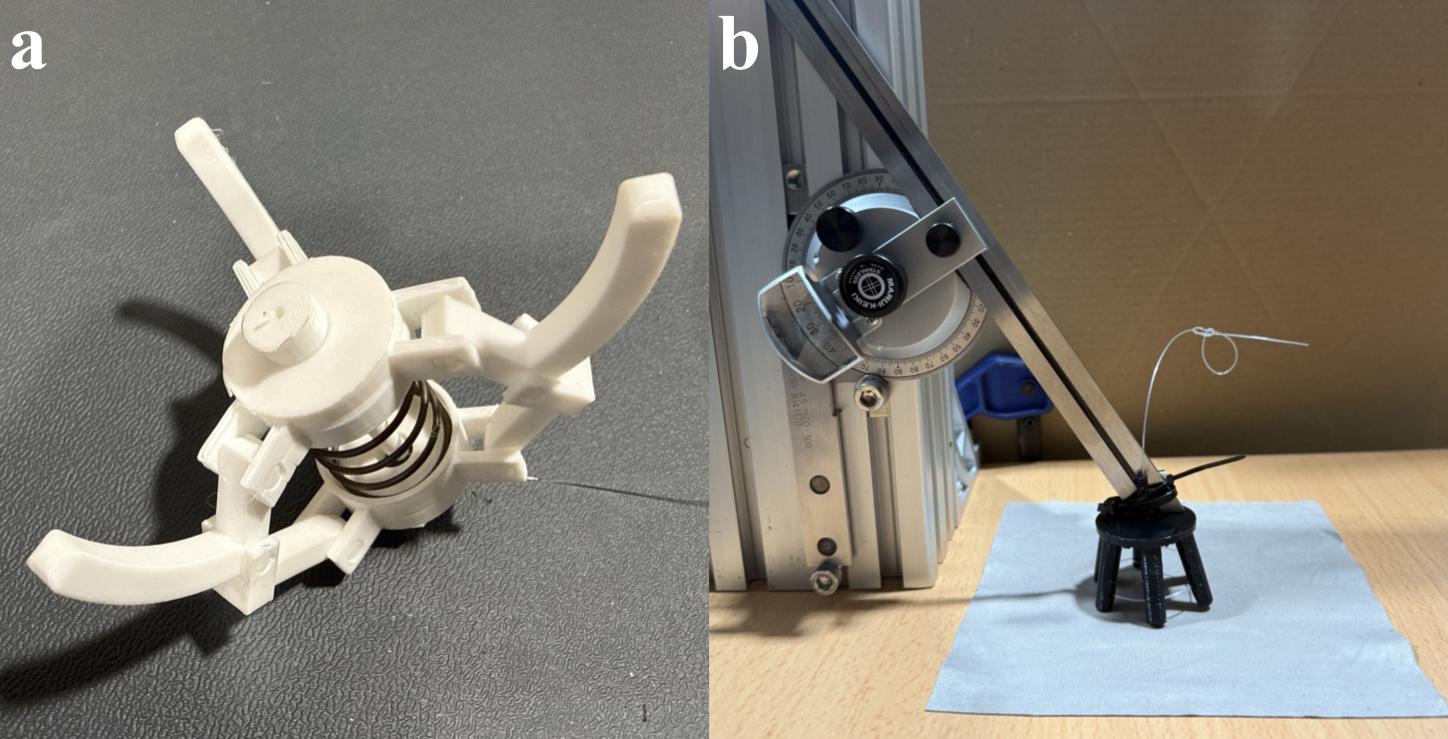}}
\caption{a) Rigid spring-loaded gripper with $N = 3$ fingers. b) Test rig for pick-and-place testing.}
\label{fig:test_setup}
\vspace{-6mm}
\end{figure}

The number of actuating gripper fingers, $N$,
and the angle, $\Phi$, of the fingers relative to the central axis of the gripper are also investigated. Seven grippers with $N \in \{2,3,4,5,6,7,8\}$ fingers at an angle of $\Phi = 20$ degrees, and four grippers with $N = 6$ fingers at angles of  $\Phi \in \{0, 20, 40, 60\} $ degrees were investigated. The tension required to obtain contact between all gripper fingers, the effective areas of contact for the open and closed gripper orientations, and the time for the gripper fingers to rebound from fully closed to fully open are presented in \cref{tab:finger_number_force}. The results for a 3-fingered rigid link, spring-loaded gripper (\cref{fig:test_setup}a) are also included to provide a baseline for comparison. The effective areas of contact were calculated to find the operational limits of each gripper, i.e.,~to find the smallest and largest object or feature that each gripper can grasp. The results for the mean rebound time indicate that as the number of contact points between the tendons and fingers increases, the friction increases, resulting in higher mean rebound times. The grippers with 7 and 8 fingers never returned to a fully relaxed state, indicating that $N = 6$ is the upper limit for $N$ at this scale. 


\begin{table}[ht]
\centering
\caption{Mean actuation force required to fully close the gripper, areas of contact, and time to rebound as a function of the number of soft fingers, $N$, and angle of soft fingers, $\Phi$.}
\label{tab:finger_number_force}
\renewcommand{\arraystretch}{1.2}
\begin{tabular}{c c c c c c}
\toprule
\makecell{Gripper} &
\makecell{Area of \\ Contact,\\Gripper \\Open\\(mm\textsuperscript{2})} &
\makecell{Area of \\ Contact,\\Gripper \\Closed\\(mm\textsuperscript{2})} &
\makecell{Mean\\Tension \\ (N)} &
\makecell{Mean\\Rebound\\Time \\ (s)} \\
\midrule
N = 2 & 1104.5 & 0.0   & $6.93 \pm 0.47$   & $5.202 \pm 1.010$ \\
N = 3 & 1104.5 & 8.0   & $9.52 \pm 0.78$   & $5.718 \pm 1.370$ \\
N = 4 & 1104.5 & 27.3  & $10.89 \pm 0.84$  & $7.574 \pm 0.757$ \\
N = 5 & 1104.5 & 52.8  & $11.85 \pm 1.45$  & $7.448 \pm 1.660$ \\
N = 6 & 1104.5 & 86.6  & $14.25 \pm 1.04$  & $10.408 \pm 3.057$ \\
N = 7 & 1104.5 & 124.7 & $14.66 \pm 1.20$  & $>60$ \\
N = 8 & 1104.5 & 167.4 & $19.45 \pm 1.20$  & $>60$ \\
\midrule
$\Phi = 0^\circ$  & 339.8  & 86.6  & $4.75 \pm 0.88$   & $7.934 \pm 0.538$ \\
$\Phi = 20^\circ$ & 1104.5 & 86.6  & $9.99 \pm 1.04$   & $10.408 \pm 3.057$ \\
$\Phi = 40^\circ$ & 2156.5 & 86.6  & $17.28 \pm 0.47$  & $16.556 \pm 3.586$ \\
$\Phi = 60^\circ$ & 3097.5 & 86.6  & $20.97 \pm 1.24$  & $17.880 \pm 2.087$ \\
\midrule
Rigid & 6607.2 & 7.8 & $11.8 \pm 187$ & $0.304 \pm 0.06$ \\
\bottomrule
\end{tabular}
\end{table}

\begin{figure}
\centerline{\includegraphics[scale=.1]{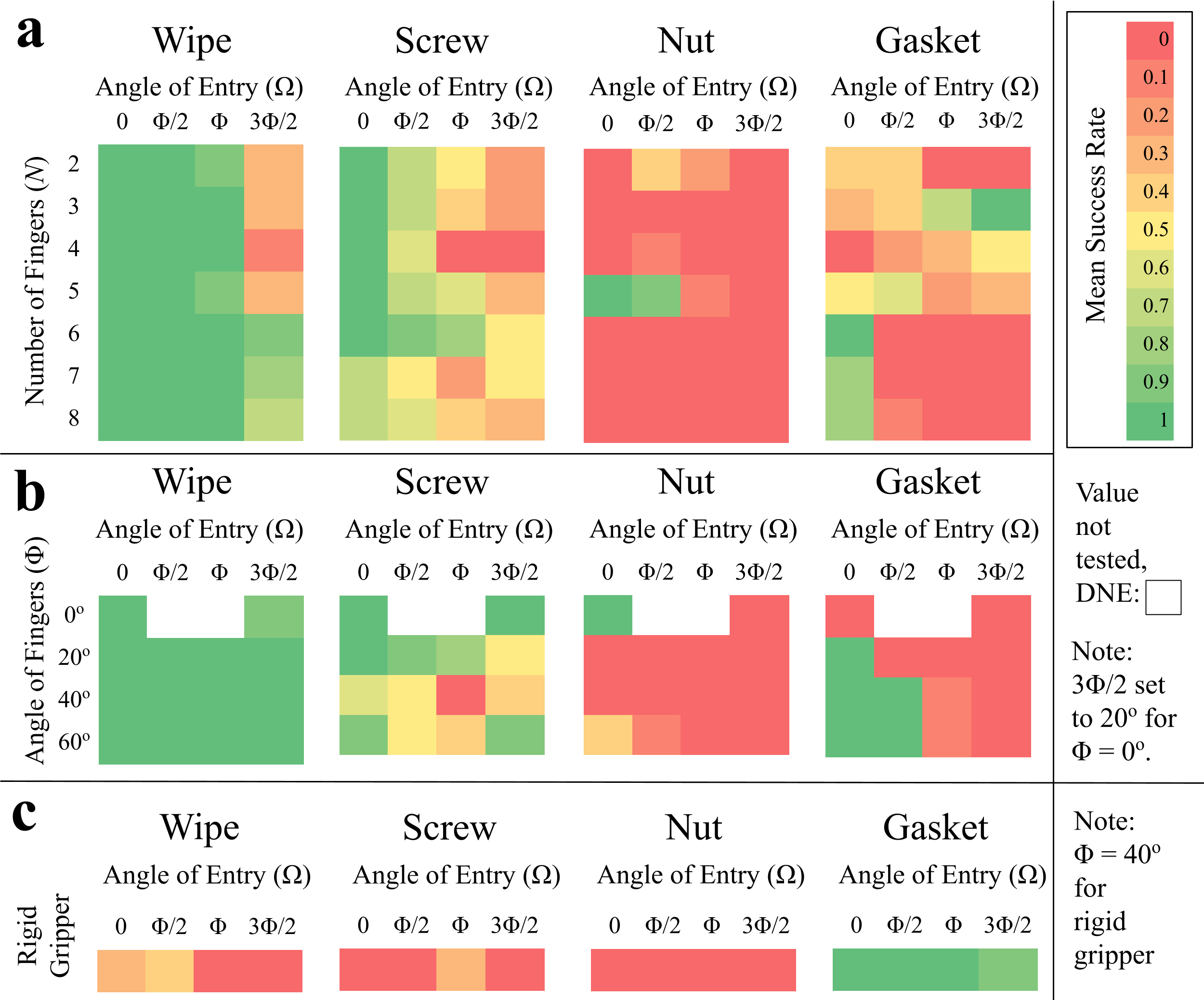}}
    \caption{Pick and place test results for four test objects, organized by a) Number of Fingers ($N$), b) Angle of Fingers ($\Phi$), and c) 3-fingered rigid gripper.}
    \label{fig:pickplace}
\vspace{-6mm}
\end{figure}

Pick-and-place testing was conducted for each soft gripper at 20\% infill and the rigid gripper for comparison. The experimental test bench set-up is shown in \cref{fig:test_setup}a and the results are presented in \cref{fig:pickplace}. In the initial prototyping stage, the gripper design was iterated with priority given to reliable wipe grasping, but to find the operational limits and versatility of the design, three additional objects are tested. The test objects selected are some which are used and/or found in intervention environments and vary in size and shape: a 140x140 mm\textsuperscript{2} beryllium testing wipe, an M6x10 screw, an M5 nut, and a rubber gasket with a diameter of 38 mm. During each test, the gripper enters the test space from a height of 10 cm and at varying angles $\Omega$. The angle $\Omega$ is measured relative to a vector normal to the ground (\cref{fig:soft_gripper}) and is evaluated at 0, $\frac{\Phi}{2}$, $\Phi$, and $\frac{3\Phi}{2}$ degrees. The gripper is oriented in the worst-case position, such that only one finger makes contact with the surface. The gripper is closed around the object and moved back to a height of 10 cm at a speed of approximately 10 mm/s to test the stability of the grasping. To randomize the position and location of the object, the object is then dropped from the gripper at a height of 1 cm. 

The results indicate that all iterations of the soft gripper design are acceptable for grasping a wipe at entry angles of $\Omega = 0$ and $\Omega = \frac{\Phi}{2}$, which remains the primary goal of the intervention. The most successful combination of parameters is shown in \cref{fig:soft_gripper}, where $N = 6$ and $\Phi = 20^\circ$. Further, since the compliant grippers can be pushed into a surface without risk, the mean success rate only decreases when the angle of entry becomes equal to or greater than the finger angle $\Phi$. In contrast, the traditional rigid gripper had a less than 50\% success rate for all entry angles and was 65\% less successful than its 3-fingered soft gripper counterpart. While more successful than all soft grippers at grasping the gasket, the rigid gripper is far less successful at grasping the wipe and less versatile across all object types. From qualitative observation, the most common failure mode for grippers with $N < 5$ fingers was poor object-tendon contact and lateral movement of the finger at which the tendon terminated, i.e.,~the tied finger. For grippers where $N > 5$, this effect was mitigated by the additional points of object-finger contact, but the smaller effective contact areas led to poor performance when grasping the smaller objects. The results for the nut highlight these phenomena, with the $N < 5$ grippers failing due to lateral movement of the tied finger and the $N > 5$ grippers failing due to the small size of the nut. The $N = 5$ gripper was the most successful. For a large object like the gasket, the grippers with more fingers ($N > 5$) perform better than their counterparts at normal entry angles, but as $\Omega$ increases, the large gaps between the grippers with fewer fingers ($N < 6$) become advantageous. Regarding finger angle, the gripper with a $20^\circ$ finger angle had the highest mean success rate. Grippers with larger finger angles performed better only in grasping the gasket, since their effective contact areas exceeded the surface area of the gasket. In contrast, due to the further travel distance and higher force required to actuate these fingers to the closed position, the force exerted on the small test objects by the approaching fingers was too high for multiple trials and pushed the objects outside of the grasp of the gripper. Limits on the speed of actuation would be required to mitigate this.



\section{Kinematic model}\label{sec:kinematic_model}

To ensure real-time performance and simple interfacing with the motors of the robot, a closed-form kinematic model is presented in this section. This model maps tendon displacements \cite{rao2021}, a property that can be read directly from the motor encoders, to robot curvature profiles.

\subsection{Actuation map}

We seek a mapping between the tendon travel distance $\boldsymbol{l} \in \mathbb{R}^3$, the bending vector $\boldsymbol{q} \in \mathbb{R}^{2}$, and robot pose $\boldsymbol{T} \in SE(3)$ at any point along the robot backbone.

The bending vector $\boldsymbol{q} = [q_x, q_y]^T$ is given as a function of $\boldsymbol{l} = [l_1, l_2, l_3]^T$:
\begin{equation}
\label{bending_vector}
    \boldsymbol{q} = \mathbf{B}\boldsymbol{l}, \quad \mathbf{B} = 
    \begin{bmatrix}
        \cos(\psi_1) & \cos(\psi_2) & \cos(\psi_3)\\
        \sin(\psi_1) & \sin(\psi_2) & \sin(\psi_3)
    \end{bmatrix},
\end{equation}
with tendon $i$ at a fixed azimuth $\psi_i = (i-1)\frac{2\pi}{3}$. The bending vector amplitude and direction are
\begin{equation}
\label{bending_vector_polar}
    \rho = ||\boldsymbol{q}|| = \sqrt{q_x^2+q_y^2}, \quad
    \phi = \text{atan} 2(q_y, q_x).
\end{equation}

\subsection{Curvature profile}

Unlike constant-curvature continuum robots, the tapered geometry of CTAM may result in spatially varying curvature profiles. The backbone tapers axis-symmetrically such that the cross-sectional circular radius $r$ decreases linearly with respect to the backbone length parameter $s$:
\begin{equation}
\label{backbone_radius}
    r(s) = r_0 + (r_1 - r_0)\frac{s}{L} = r_0+\alpha s,
\end{equation}
where $r_0$ and $r_1$ denote the cross-section radii at the base and the tip, respectively, and $\alpha$ is the taper angle. To maintain a slender robot geometry, the tendon routing also follows a linear taper along $s$:
\begin{equation}
\label{tendon_offset}
    d(s) = d_0 + (d_1 - d_0)\frac{s}{L} = d_0 + \beta s,
\end{equation}
where each tendon is located at a radial offset $d(s)$ from the neutral axis, and $d_0$, $d_1$ and $\beta$ are defined analogous to $r_0$, $r_1$ and $\alpha$. The net bending moment is
\begin{equation}
    M(s) = \tau d(s) = M_0 \frac{d(s)}{d_0},
\end{equation}
where $\tau \in \mathbb{R}^3$ is the tendon tension and $M_0 = \tau d_0$ is the moment that would be applied by the base moment arm. Under the Euler-Bernoulli assumption \cite{murray2017}, the curvature is given by
\begin{equation}
\label{kappa}
    \kappa (s) = \frac{M_0d(s)}{E\frac{\pi}{4}(r(s)^4 - r_{in}^4)d_0} = \kappa_{0}w(s),
\end{equation}
where $E$ is the Young's modulus of the backbone, $\frac{\pi}{4}(r(s)^4 - r_{in}^4)$ is the second moment of area for a hollow circular cross-section, $r_{in}$ is the constant inner radius of the backbone, $\kappa_0 = \frac{4M_0}{\pi E(r_0^4 - r_{in}^4)}$ is the reference curvature at the base, and $w(s) = \frac{d(s)(r_0^4 - r_{in}^4)}{d_0(r(s)^4-r_{in}^4)}$ is the local curvature weight.

Since the Euler-Bernoulli model assumes homogeneous, isotropic, and time-independent linear elasticity, we propose an alternative single-parameter affine local curvature weight that can be fitted to real data samples from motion tracking:
\begin{equation}
    w(s) = 1 + \gamma (1 - \frac{2s}{L}),
\end{equation}
where $\gamma$ dictates the shape. Note that $\gamma = 0$ reduces the model to the constant curvature model.

The cumulative bending angle is
\begin{subequations}
    \begin{equation}
    \label{theta}
        \theta (s) = \int^s_0 \kappa (\sigma)d\sigma = \kappa_0 W(s),
    \end{equation}
    \begin{equation}
        W(s) = \int^s_0 w (\sigma)d\sigma.
    \end{equation}
\end{subequations}
As both the presented alternatives of $w(s)$ admit a closed-form of $W(s)$, the cumulative bending angle $\theta$ at any point along the backbone can be also obtained in closed-form.

The base curvature $\kappa_0$ is constrained by the tendon displacements through the kinematic equivalence between the effective in-plane tendon displacement and the bending vector magnitude:
\begin{equation}
    \rho = \int^L_0d(s)\kappa(s)ds = \kappa_0d_0LC,
\end{equation}
where $C$ also has a closed form.
Now, the base curvature is given by
\begin{equation}
\label{kappa_0}
    \kappa_0 = \frac{\rho}{d_0LC},
\end{equation}
such that the mapping from tendon displacement to the cumulative bending magnitude follows $\boldsymbol{l} \rightarrow \rho \rightarrow \kappa_0 \rightarrow \theta(s)$ for any $s$, and the direction is found directly through $\boldsymbol{l} \rightarrow \phi$ independently of $s$.

\subsection{Forward kinematics}

Given the curvature profile, the robot pose can now be obtained in closed form. Under the assumption of pure backbone bending without torsion, that is, constant $\phi$ over $s$, the orientation of the centerline of the backbone is $\boldsymbol{R}(s) = \boldsymbol{R}_z(\phi) \boldsymbol{R}_y(\theta(s)) \boldsymbol{R}_z(-\phi)$, or equivalently,
\begin{equation}
    \boldsymbol{R}(s) =
    \begin{bmatrix}
        1 - (1 - c\theta)c\phi^2 & -(1 - c\theta)s\phi c\phi & s\theta c\phi \\
        -(1 - c\theta)s\phi c\phi & 1- (1 - c\theta)s\phi^2 & s\theta s\phi \\
        -s\theta c\phi & -s\theta s\phi & c\theta
    \end{bmatrix},
\end{equation}
where $c\cdot = \cos(\cdot)$, $s\cdot = \sin(\cdot)$ and $\theta = \theta(s)$. The centerline position $\boldsymbol{p}(s)$ is found by integrating along the arc following the unit tangent $\hat{\boldsymbol{t}} = \boldsymbol{R}(s)\hat{\boldsymbol{z}} = [s\theta c\phi, s\theta s\phi, c\theta]^T$:
\begin{equation}
    \boldsymbol{p}(s) = \int_0^s\hat{\boldsymbol{t}}(\sigma)d\sigma =
    \begin{bmatrix}
        c\phi \int_0^s \sin(\theta(\sigma))d\sigma\\
        s\phi \int_0^s \sin(\theta(\sigma))d\sigma\\
        \int_0^s \cos(\theta(\sigma))d\sigma
    \end{bmatrix}.
\end{equation}
The homogeneous centerline point transform is 
\begin{equation}
    \boldsymbol{T}(s) = \begin{bmatrix}
        \boldsymbol{R}(s) & \boldsymbol{p}(s)\\
        \boldsymbol{0}^T & 1
    \end{bmatrix}.
\end{equation}

\subsection{Differential kinematics/Jacobian}

For task-space control, a differential mapping between tendon motions and end-effector motion is required. We seek a Jacobian $\boldsymbol{J} \in \mathbb{R}^{6 \times 3}$ with $\dot{\boldsymbol{x}} = [\dot{\boldsymbol{p}}^T, \boldsymbol{\omega}^T]^T$, where $\boldsymbol{\omega}$ is the angular velocity of the centerline given by $s$. The map is decomposed into
\begin{equation}
\label{jacobian}
    \boldsymbol{J} = \underbrace{\begin{bmatrix}
        \frac{\partial \boldsymbol{p}}{\partial\kappa_0} & \frac{\partial \boldsymbol{p}}{\partial\phi} \\
        \frac{\partial \boldsymbol{\omega}}{\partial\kappa_0} & \frac{\partial \boldsymbol{\omega}}{\partial\phi}
    \end{bmatrix}}_{\boldsymbol{J}_{pM}}
    \underbrace{\begin{bmatrix}
        \frac{\partial \kappa_0}{\partial \boldsymbol{q}} & \frac{\partial \phi}{\partial \boldsymbol{q}} \\
    \end{bmatrix}}_{\boldsymbol{M}(\boldsymbol{q})}
    \underbrace{\frac{\partial \boldsymbol{q}}{\partial \boldsymbol{l}}}_{\boldsymbol{B}}.
\end{equation}
From \eqref{bending_vector}, the relation $\frac{\partial \boldsymbol{q}}{\partial \boldsymbol{l}} = \boldsymbol{B}$ is known. From \eqref{bending_vector_polar} and \eqref{kappa_0}, we have
\begin{equation}
\label{dkappa0phi_dq}
    \frac{\partial \kappa_0}{\partial \boldsymbol{q}} =
    \begin{bmatrix}
        \frac{q_x}{d_0LC\rho}\\
        \frac{q_y}{d_0LC\rho}
    \end{bmatrix},\quad
    \frac{\partial \phi}{\partial \boldsymbol{q}} =
    \begin{bmatrix}
        -\frac{q_y}{\rho^2} \\
        \frac{q_x}{\rho^2}
    \end{bmatrix}.
\end{equation}
Note that $C$ depends only on fixed geometry and not $\boldsymbol{l}$. For the pose block in \eqref{jacobian}, the linear velocity parts are
\begin{subequations}
    \begin{equation}
        \frac{\partial \boldsymbol{p}}{\partial \kappa_0} =
        \begin{bmatrix}
            c\phi \int_0^s W(\sigma)\sin(\theta(\sigma))d\sigma\\
        s\phi \int_0^s W(\sigma)\sin(\theta(\sigma))d\sigma\\
        -\int_0^s W(\sigma)\cos(\theta(\sigma))d\sigma
        \end{bmatrix},
    \end{equation}
    \begin{equation}
        \frac{\partial \boldsymbol{p}}{\partial \phi} =\int_0^s \sin(\theta(\sigma))d\sigma
        \begin{bmatrix}
            -s\phi\\c\phi\\0
        \end{bmatrix}.
    \end{equation}
\end{subequations}
The angular velocity parts are
\begin{equation}
    \frac{\partial \boldsymbol{\omega}}{\partial \kappa_0} = W(s)
    \begin{bmatrix}
        -s\phi\\c\phi\\0
    \end{bmatrix}, \quad \frac{\partial \boldsymbol{\omega}}{\partial \phi} =
    \begin{bmatrix}
        -c\phi \sin(\theta(s))\\
        s\phi \sin(\theta(s))\\
        1 - \cos (\theta(s))
    \end{bmatrix}
\end{equation}

In order to avoid zero configuration singularities, i.e.,~$\rho = 0$, we divide the actuation-to-task mapping into a linear and a nonlinear part and define the following base curvature vector:
\begin{equation}
\label{kappa_*}
    \boldsymbol{\kappa}_* = \kappa_0\begin{bmatrix}
        \cos(\phi)\\
        \sin(\phi)
    \end{bmatrix}
    =\boldsymbol{K}_\kappa
    \boldsymbol{l}, \quad \boldsymbol{K}_\kappa = \frac{1}{d_0LC}\boldsymbol{B}.
\end{equation}
Note that the mapping from actuation to the base curvature vector is linear in $\boldsymbol{l}$. The pose $\boldsymbol{x}$ is a smooth function of $\boldsymbol{\kappa}_*$, and the Jacobian
\begin{equation}
\label{jacobian_reparametrized}
    \boldsymbol{J} = \boldsymbol{J}_{x\kappa}\boldsymbol{K}_\kappa, \quad \boldsymbol{J}_{x\kappa} =  \frac{\partial \boldsymbol{x}}{\partial \boldsymbol{\kappa}_*}
\end{equation}
is well-defined even for $\boldsymbol{\kappa}_* = \boldsymbol{0}$ when the Jacobian rank drops from $2$ to $1$, despite the division by $\rho$ in \eqref{dkappa0phi_dq}.

\section{Control}\label{sec:control}

In this section, we present a closed-loop inverse-kinematics (CLIK) task-space controller that generates minimum-norm tendon motions while preserving an additional common-mode degree of freedom that does not affect the robot curvature configuration. The controller exploits the linear mapping between tendon motions and the base curvature vector introduced in \cref{sec:kinematic_model}.

The controller is organized as a two-loop architecture, following the re-parameterized structure in \eqref{jacobian_reparametrized}. The inner loop takes the desired base curvature vector rate $\dot{\boldsymbol{\kappa}}_{*d}$ from the outer loop and provides a commanded tendon velocity:
\begin{equation}
    \dot{\boldsymbol{l}} = \boldsymbol{K}_{\kappa}^\dagger \dot{\boldsymbol{\kappa}}_{*d} + \nu \boldsymbol{1},
\end{equation}
where $\boldsymbol{K}_{\kappa}^\dagger$ denotes the Moore-Penrose pseudo-inverse of the full-rank $\boldsymbol{K}_{\kappa}$, which is constant and can be computed offline. Note that $\boldsymbol{K}_{\kappa}\dot{\boldsymbol{l}} = \dot{\boldsymbol{\kappa}}_{*d} \forall \nu \in \mathbb{R}$, because the null space of $\boldsymbol{B}$ is given by $\text{span}\left\{[1, 1, 1]^T\right\}$, which implies $\boldsymbol{K}_{\kappa} \boldsymbol{1} = \boldsymbol{0}$. This allows pretension control to be decoupled from the task-space control by construction. This provides an extra degree of freedom in the common-mode direction $\nu$ for controlling the axial preload of the robot without affecting its curvature configuration. Consequently, the robot can exert greater interaction forces when required, e.g.,~during wall swabs.

The outer loop maps task-space tracking objectives to desired curvature-vector motions. It consists of a singularity-robust CLIK law:
\begin{equation}
    \dot{\boldsymbol{\kappa}}_{*d} = \boldsymbol{J}_{x\kappa}^T(\boldsymbol{J}_{x\kappa}\boldsymbol{J}_{x\kappa}^T + \lambda^2 \boldsymbol{I}_3)^{-1}(\dot{\boldsymbol{p}}_d + \boldsymbol{\Lambda e}),
\end{equation}
with the task gain matrix $\boldsymbol{\Lambda} = \boldsymbol{\Lambda}^T \succ 0$. The $\lambda$-damped pseudo-inverse of $\boldsymbol{J}_{x\kappa}$ provides a least-squares feasible robot motion also when $\boldsymbol{p}_d$ leaves the reachable manifold, as well as regularizing the zero-curvature singularity $\rho = 0$ where $\phi$ is undefined:
\begin{equation}
    \lambda^2 = \begin{cases}
        0, & \rho \leq \epsilon\\
        \lambda_0^2(1 - (\rho/\epsilon)^2), & \rho > \epsilon
    \end{cases}.
\end{equation}
Here, $\epsilon$ is a threshold below which singular values are considered near-singular \cite{siciliano2016}.

\section{Experimental results}\label{sec:experimental_results}

In this section, we assess the kinematic model from \cref{sec:kinematic_model}. Specifically, we evaluate the extent to which the common-mode shape-independence assumption is satisfied, and compare its effect on the robot shape to the effects of opening and closing of the gripper.

The prototype used in the experiments comprises a hollow tapered backbone with $L = 405.1$ mm, $r_0 = 7.48$ mm, $r_1 = 3.94$ mm, and $r_{in} = 2.20$ mm, following the notation in \cref{sec:kinematic_model}. The tendon offsets are $d_0 = 26.69$ mm and $d_1 = 10.72$ mm. All tendons are driven by Dynamixel XH430-W210T servo motors operated in current mode. The motion capture was performed using Vicon
Tracker 3.10.2 with Vicon Vero grayscale cameras. Visual marker triplets were affixed to three points on the prototype: One near the base, one on the outer circumference of a disc (the sixth disc from the base) of $24$ mm radius located at $107$ mm along the backbone from the base, and one at the tip. The cameras captured movements with a position resolution of $0.02$ mm. The poses of the disc and the tip are both relative to the robot base.

\begin{figure}
\centerline{\includegraphics[scale=.75]{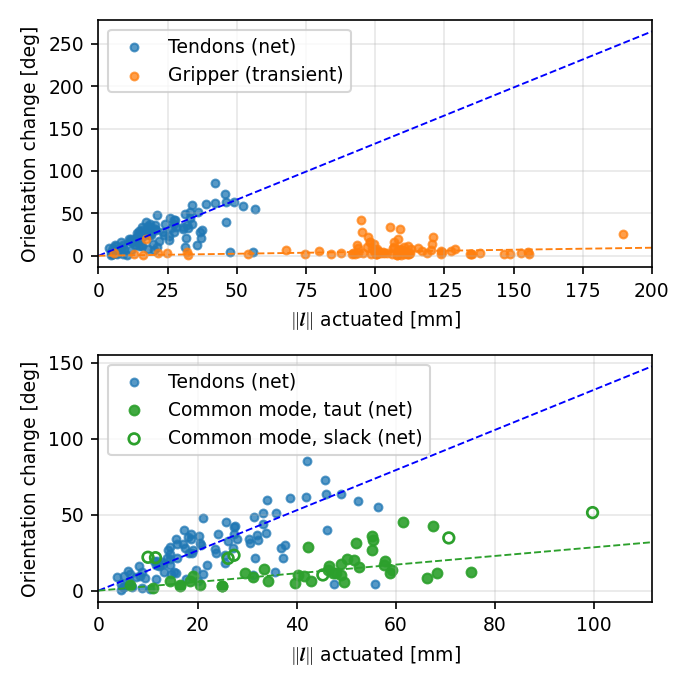}}
\caption{Tip orientation response against tendon displacement. Net bending tendon compared to gripper (top) and common-mode (bottom).}
\label{fig:scatter}
\vspace{-6mm}
\end{figure}

The kinematic model assumes that the common-mode actuation is decoupled from the kinematics. Moreover, the backbone is treated load-free, neglecting the effects of the gripper. For evaluating these assumptions, we pooled 85 gripper open/close events, 97 bending maneuvers, and 49 common-mode maneuvers from a series of motion tracking trials. \cref{fig:scatter} compares the robot tip orientation change caused by the tendon displacement changes of these events. \cref{tab:gripper_cm_sens} displays the sensitivity values, indicating a more significant kinematic coupling for the common-mode than for the gripper. Therefore, the worst-case transient response is used for the gripper, while the common-mode is represented by the permanent positioning. Furthermore, the common-mode separates tendon slack from taut, as at least one tendon moved beyond the pretensioned reference for the slacking. For these events, the kinematic decoupling is void, and motor positions no longer determine the configuration.

\begin{table}[t]
\centering
\caption{Median tip and disc orientation sensitivity per mm of actuated tendon. Percentages are relative to the bending tendons.}
\label{tab:gripper_cm_sens}
\begin{tabular}{lccc}
\toprule
Actuation class & Events & Tip [deg/mm] & Disc [deg/mm]\\
\midrule
Bending tendons & $97$ & $1.322$ & $0.698$\\
Gripper (transient) & $85$ & $0.047$ ($3.6\%$) & $0.022$ ($3.2\%$)\\
Common mode, taut & $42$ & $0.285$ ($21.6\%$) & $0.116$ ($16.6\%$)\\
Common mode, slack & $7$ & $0.881$ ($61\%$) & $0.224$ ($32\%$)\\
\bottomrule
\end{tabular}
 \end{table}

To compensate for compliance effects, manufacturing tolerances, and deviations from the idealized Euler-Bernoulli assumptions, the kinematic model was scaled by a scalar $k_{cal}$ that was found through a one-dimensional least-square fit $k_{cal} = \argmin_{k \in \mathbb{R}} \sqrt{\frac{1}{N}\sum_{i=1}^N(f(k)^2)}$ with
\begin{equation}
    f(k) = \theta_{tip,vicon} - ||k\boldsymbol{K}_{\kappa}\boldsymbol{l}_{enc} + \boldsymbol{\kappa}_{*,vicon}||W(L),
\end{equation}
where $\theta_{tip,vicon}$ is the measured tip position. Furthermore, $\boldsymbol{\kappa}_{*,vicon}$ is the estimated initial curvature vector based on the first samples per trial to account for the robot not being perfectly stiff, owing to the backbone viscoplastic effects of repeated bending and loading. The values in $\boldsymbol{l}_{enc}$ are read from the motor encoder and converted to tendon displacements.

\begin{figure}
\centerline{\includegraphics[scale=.75]{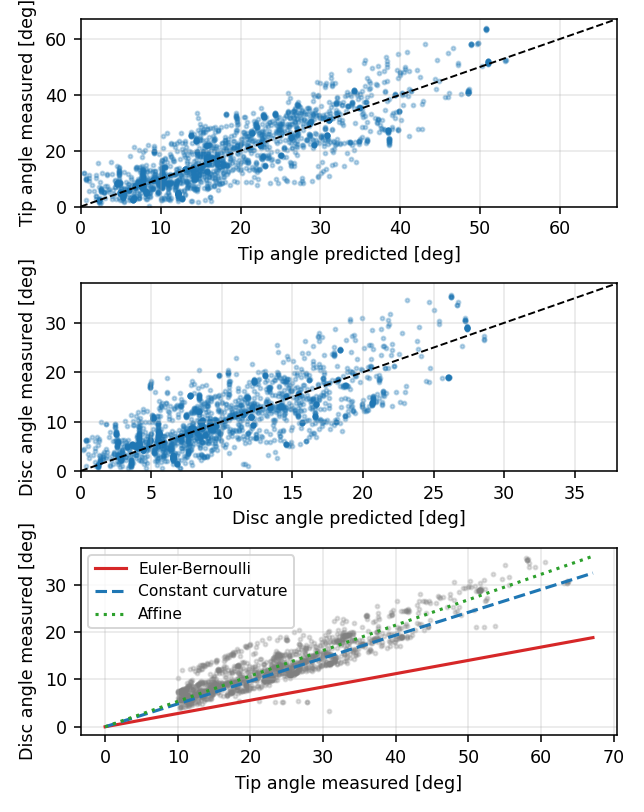}}
\caption{Orientation measurements against affine curvature model predictions for the tip (top) and sixth disc (middle), and calibration-free shape comparison (bottom).}
\label{fig:model_assessment}
\vspace{-4mm}
\end{figure}

\cref{fig:model_assessment} (top and middle) display the comparison between the affine curvature model orientation predictions and the measurements. Note that $\boldsymbol{K}_{\kappa}$ in \eqref{kappa_*} was scaled by a factor of $k_{cal}$. As there were seven separate recorded trials, $k_{cal}$ was found with a grid search for each trial, with $k_{cal} = 0.602 \pm 0.069$. The shape variable of the affine curvature function was similarly found to be $\gamma = 0.213$. The sixth disc comparison used the same $k_{cal}$ value found only from tip measurements. The bottom plot shows a shape comparison between the Euler-Bernoulli, affine and constant curvature models. The resulting root-mean-square prediction error over all trials was $5.6$ deg for the tip angle. For the disc angle, a hold-out trial was used for validation, resulting in an error of $7.3$ deg for the Euler-Bernoulli model, $2.9$ deg for the constant curvature, and $2.5$ deg for the affine curvature shape.

To assess the stiffening effect of the common-mode pretension, the robot tip was pushed with known horizontal forces while the tip displacements were recorded at different common-mode tendon displacements, using the motion capture system together with an RS Pro 111-3689 force gauge. \cref{tab:stiffness_pretension} shows the mean common-mode tendon travel against tip deflections while pushing the tip with a fixed force of 0.20~N. The data strongly support the claim that common-mode pretension stiffens the robot against external forces.

\begin{table}[t]
\centering
\caption{Tip deflection and effective stiffness under a 0.20~N tip push, versus cumulative common-mode tendon travel.}
\label{tab:stiffness_pretension}
\begin{tabular}{lcccc}
\toprule
travel [mm] & mean travel [mm] & tip deflection [mm] & $k_{\mathrm{eff}}$ [N/m]\\
\midrule
$0$               & $0.0$  & $50.2 \pm 4.8$ & $3.9 \pm 0.4$\\
$0$--$15$         & $10.9$ & $16.3 \pm 0.3$ & $12.0 \pm 0.2$\\
$15$--$22$        & $18.0$ & $13.5 \pm 2.8$ & $15.1 \pm 3.0$\\
$22$--$34$        & $28.1$ & $11.7 \pm 0.3$ & $16.7 \pm 0.4$\\
\bottomrule
\end{tabular}
\vspace{-4mm}
\end{table}

\section{Conclusion}\label{sec:conclusion}

This work presented a novel quadruped soft tail solution for contamination surveys in cluttered environments. The soft tail is tendon actuated and features a hollow backbone through which an additional tendon can traverse through to actuate a soft gripper designed for picking up tissues from flat surfaces. We present a closed-form kinematic model allowing for varying curvature profiles. Based on these models, we have developed a task-space controller augmented with shape-independent tendon pretension control. This controller enables accurate tissue pick-and-place and compliant wall swabbing tasks for teleoperated beryllium contamination surveys.

Experimental results demonstrated that gripper actuation has a negligible effect on robot shape, while common-mode tendon actuation provides a useful mechanism for stiffness modulation and preload control, despite a stronger kinematic coupling. The experiments further demonstrated that the affine curvature model provided the best approximation of the backbone deformation behavior, and serves as a practical basis for real-time task-space control.

Future work will include conducting user operation studies of the intervention.


\vspace{-4mm}

\phantomsection
\section*{APPENDIX} \label{appendix}

Video link: \mbox{\url{https://youtu.be/st8n9rzAASI}}.

\bibliographystyle{IEEEtran}
\bibliography{references.bib}{}

\end{document}